\documentclass{article}

\usepackage[preprint]{neurips_2026}
\usepackage[utf8]{inputenc}
\usepackage[T1]{fontenc}
\usepackage{hyperref}
\usepackage{url}
\usepackage{booktabs}
\usepackage{multirow}
\usepackage{amsfonts}
\usepackage{amsmath}
\usepackage{amssymb}
\usepackage{nicefrac}
\usepackage{microtype}
\usepackage{xcolor}
\usepackage{graphicx}
\usepackage{enumitem}
\usepackage{listings}

\definecolor{codekw}{HTML}{6A1B9A}   
\definecolor{codestr}{HTML}{1B5E20}  
\definecolor{codecom}{HTML}{616161}  
\definecolor{codebg}{HTML}{F7F7F7}   
\lstdefinestyle{pyfem}{
  language=Python,
  basicstyle=\ttfamily\scriptsize,
  keywordstyle=\color{codekw}\bfseries,
  stringstyle=\color{codestr},
  commentstyle=\color{codecom}\itshape,
  backgroundcolor=\color{codebg},
  showstringspaces=false,
  breaklines=true,
  columns=fullflexible,
  keepspaces=true,
  xleftmargin=4pt,
  xrightmargin=4pt,
  aboveskip=2pt,
  belowskip=2pt,
  morekeywords={TrialFunction,TestFunction,inner,grad,dx,
                functionspace,create_square_mesh,LinearProblem},
}
\lstdefinestyle{promptcard}{
  basicstyle=\ttfamily\scriptsize,
  keywordstyle=\color{codekw}\bfseries,
  backgroundcolor=\color{codebg},
  showstringspaces=false,
  breaklines=true,
  columns=fullflexible,
  keepspaces=true,
  xleftmargin=4pt,
  xrightmargin=4pt,
  aboveskip=3pt,
  belowskip=3pt,
  morekeywords={DIAGNOSIS,METHOD},
}

\title{AutoPDE: Reliable Agentic PDE Solving via Explicitly Represented Solver Strategies}

\author{%
Huanshuo Dong\textsuperscript{1,*} \quad
Keyao Zhang\textsuperscript{1,*} \quad
Hong Wang\textsuperscript{1} \quad
Zhezheng Hao\textsuperscript{2} \\
Zhiwei Zhuang\textsuperscript{1} \quad
Ziyan Liu\textsuperscript{1} \quad
Jiacong Wang\textsuperscript{3} \quad
Gengyuan Liu\textsuperscript{4} \quad
Xin Jin\textsuperscript{5,\ensuremath{\dagger}} \\
\textsuperscript{1}University of Science and Technology of China \\
\textsuperscript{2}Zhejiang University \quad
\textsuperscript{3}University of the Chinese Academy of Sciences \\
\textsuperscript{4}Tsinghua University \quad
\textsuperscript{5}Eastern Institute of Technology, Ningbo \\
\textsuperscript{*}Equal contribution \quad
\textsuperscript{\ensuremath{\dagger}}Corresponding author \\
\texttt{bingo000@mail.ustc.edu.cn} \quad
\texttt{jinxin@eitech.edu.cn}
}

\begin{document}

\maketitle
\begin{abstract}
Numerical solvers for partial differential equations (PDEs) are core computational tools in science and engineering.
Building reliable PDE solvers requires not only executable code, but a \emph{numerical solver strategy}, a set of decisions about discretization, stabilization, solver configuration, and resolution control, that matches the PDE structure.
Recent LLM-based coding agents have begun to reduce the programming burden by generating and debugging solver implementations.
However, they typically move directly from a PDE problem to solver code, leaving the solver strategy implicit in implementation details.
Feedback from a failed solve is therefore routed back to code edits rather than to the underlying strategy, so numerical decisions remain hard to check before code is generated and hard to revise using numerical evidence when it fails.
To address this limitation, we propose AutoPDE, a code agent that maintains the solver strategy as an \emph{explicitly represented} object throughout the solving process: an independent, inspectable object that is built before any code is written and can be revised, using numerical evidence, whenever a solve fails.
AutoPDE builds and maintains this object in three stages, all drawing from a library of reusable PDE-solving skills: \emph{PDE analysis} identifies the equation type and algebraic structure; \emph{numerical method selection} chooses a numerical method that matches the analysis result and commits to a discretization, stabilization, and linear solver accordingly; and \emph{adaptive tuning} runs low-cost pilot solves to calibrate resolution and tolerances under the prescribed accuracy and runtime budget.
We evaluate AutoPDE on the PDE Agent Bench, where experimental results show that AutoPDE achieves a pass rate of $54.5\%$, improving over the strongest baseline by $14.2$ percentage points.

\end{abstract}

\section{Introduction}


Numerical solvers for partial differential equations (PDEs) are core computational tools in science and engineering, playing a central role in the simulation of fluid dynamics, heat transfer, elasticity, wave propagation, and other physical problems. Building reliable PDE solvers is not only about writing executable code that runs without errors. A more fundamental challenge is designing a numerical solver strategy that matches the PDE structure. This strategy consists of the choice of discretization, stabilization, solver configuration, and resolution control. Without the right strategy, even a program that runs can be mathematically inappropriate or computationally inefficient.


Recent LLM-based coding agents have begun to address the implementation side of PDE solver construction. By interacting with a compute environment, these agents can generate code, execute it, observe failures, and iteratively edit the implementation. General-purpose systems such as SWE-agent~\citep{yang2024sweagent} and OpenHands~\citep{wang2024openhands} demonstrate the strength of this paradigm on software engineering benchmarks. PDE-specific work such as CodePDE~\citep{li2025codepde} further shows that LLMs can generate numerical solver programs and improve them through debugging, refinement, and test-time scaling.


However, these systems remain program-centric. The solver strategy is represented only implicitly through implementation details. There is no separate artifact that records why a discretization, stabilization, solver configuration, or resolution was chosen. Even when execution results or accuracy feedback are used, the feedback is applied to generated programs rather than to the solver strategy itself. This makes numerical choices hard to inspect before code generation and hard to revise in a targeted way when solving fails. For example, applying conjugate gradient to an indefinite Helmholtz system~\citep{saad2003iterative}, or omitting SUPG stabilization for a convection-dominated problem~\citep{brooks1982streamline}, reflects a strategy-level mismatch rather than a mere coding error. Such failures are difficult to fix by local code patching alone, because the agent must revisit the underlying numerical decision rather than only the implementation.

\begin{figure}[t]
  \centering
  \includegraphics[width=\linewidth]{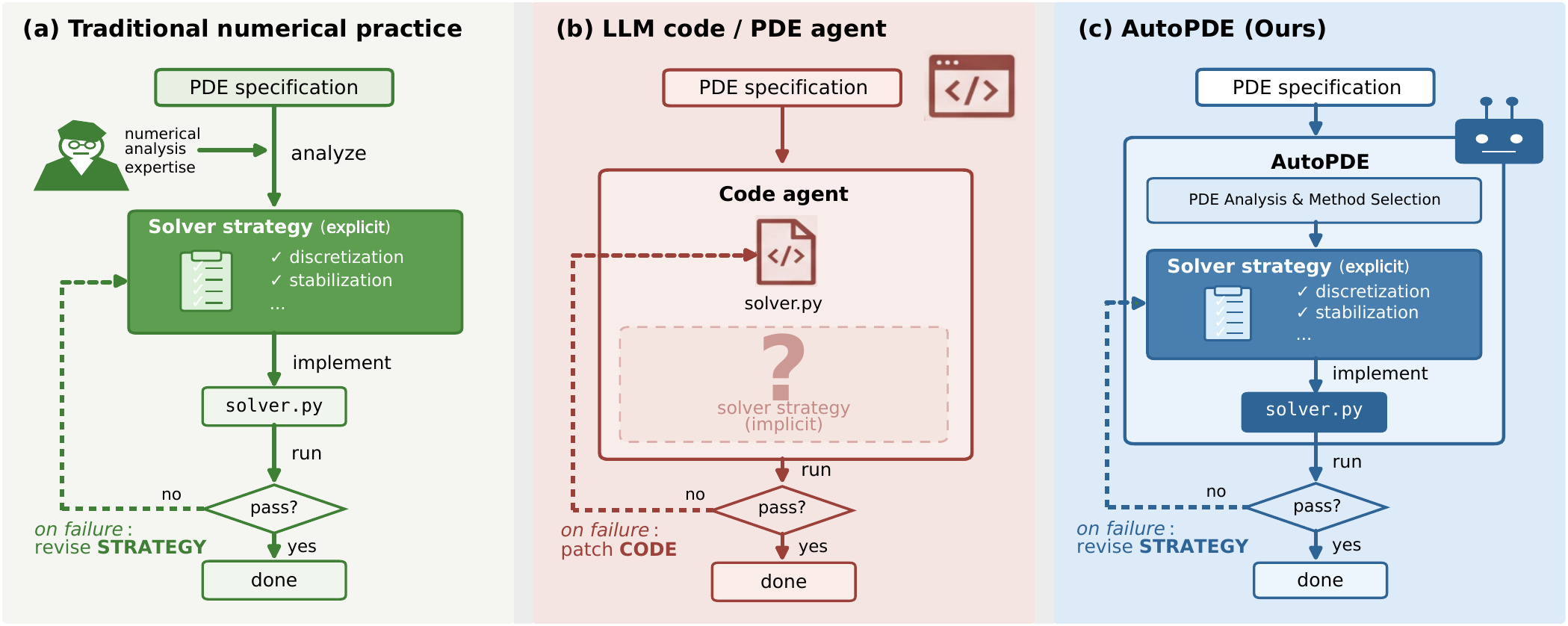}
  \caption{%
    Three ways to build a PDE solver.
    \textbf{(a) Traditional numerical practice:} a human expert applies numerical-analysis knowledge to derive an \emph{explicit} solver strategy (discretization, stabilization, solver configuration), implements it, and \emph{on failure revises the strategy} rather than only the code.
    \textbf{(b) LLM code / PDE agent:} a general code agent writes \texttt{solver.py} directly from the PDE specification; the solver strategy remains implicit inside the generated code, so when the solve fails the agent can only \emph{patch the code}, not the underlying numerical decision.
    \textbf{(c) AutoPDE (ours):} AutoPDE first runs an explicit PDE-analysis and numerical-method-selection stage, maintains the solver strategy as a first-class object alongside \texttt{solver.py}, and \emph{on failure revises the strategy} using numerical evidence, recovering the (a)-style feedback loop while keeping the full process automated.%
  }
  \label{fig:intro}
\end{figure}

To address these limitations, we propose AutoPDE, a PDE solving agent built around \emph{explicitly represented solver strategies}. Rather than writing code directly, AutoPDE first constructs a solver strategy as an independent, inspectable object, and then realizes it as executable code. The strategy is maintained throughout the solving process: numerical decisions remain visible, checkable, and revisable using numerical evidence whenever a solve fails. Figure~\ref{fig:intro} contrasts this design with traditional numerical practice, where a human expert makes the same strategy explicit by hand, and with current LLM code agents, where the strategy is left implicit inside the generated code: AutoPDE automates the former loop rather than following the latter one. AutoPDE builds and maintains this object in three stages, all drawing from a library of reusable PDE-solving skills. \emph{PDE analysis} identifies the PDE type, boundary conditions, dominant mechanisms, and algebraic structure. \emph{Numerical method selection} chooses a numerical method that matches the analysis result and commits to a variational formulation, discretization, stabilization, linear solver, and preconditioner accordingly. \emph{Adaptive tuning} runs low-cost pilot solves to measure empirical error and wall-clock time, and then calibrates resolution and solver tolerances under prescribed accuracy and runtime constraints.

Our contributions are as follows:
\begin{itemize}[leftmargin=1.5em]
  \item \textbf{Strategy-first PDE agent design.} We propose AutoPDE, which makes the solver strategy an explicit, auditable object rather than leaving numerical decisions implicit in generated code. This design makes method choices visible, checkable, and revisable throughout the solving process.
  \item \textbf{Reusable skills for solver strategy construction.} We build solver strategies from reusable PDE solving skills. PDE analysis records the equation structure, while numerical method selection composes skills for formulation, discretization, stabilization, solver selection, and preconditioner configuration.
  \item \textbf{Adaptive tuning from numerical evidence.} We introduce a pilot-solve tuning mechanism that measures empirical error and wall-clock time, then adjusts resolution and solver tolerances under prescribed accuracy and runtime constraints. On PDE Agent Bench, AutoPDE reaches an overall pass rate of $54.5\%$ and matches this pass rate on two different backbones (Claude~Opus~4.6 and GPT~5.1), improving over the strongest baseline by $14.2$ percentage points.
\end{itemize}

\section{Related Work}

\paragraph{Numerical simulation software and finite-element algorithms.}
Classical PDE software such as FEniCS~\citep{logg2012automated,alnaes2015fenics}, Firedrake~\citep{rathgeber2016firedrake}, and deal.II~\citep{bangerth2007deal} exposes the numerical pipeline as explicit choices: discretization spaces~\citep{taylor1973numerical,arnold1984stable,raviart1977mixed}, stabilization for non-coercive or convection-dominated regimes~\citep{brooks1982streamline,hughes1989new}, Krylov solvers and preconditioners~\citep{saad1986gmres,saad2003iterative,ruge1987algebraic,benzi2005numerical,elman2014finite}, and adaptivity~\citep{verfurth2013posteriori,ainsworth2000posteriori,babuvska1994p}. AutoPDE lifts this human-curated workflow into an agentic setting by making the solver strategy an explicit, revisable artifact constructed from reusable numerical-method skills.

\paragraph{Agentic software engineering.}
LLM coding agents such as SWE-agent~\citep{yang2024sweagent}, OpenHands~\citep{wang2024openhands}, and MetaGPT~\citep{hong2024metagpt} interact with compute environments, execute code, and edit implementations using feedback, often through self-debugging~\citep{chen2024selfdebug} or self-refinement~\citep{madaan2023selfrefine}. These systems are effective general software agents, but they do not explicitly represent coupled PDE decisions such as formulation, discretization, stabilization, and solver/preconditioner choice. AutoPDE inherits the interactive agentic paradigm while adding a PDE-specific strategy representation and skill library.

\paragraph{LLM-driven PDE solver generation.}
Recent PDE-specific systems formulate solver construction as code generation. CodePDE~\citep{li2025codepde} studies repeated sampling, self-debugging, refinement, and test-time scaling; PDE-SHARP~\citep{fazliani2025pdesharp} uses staged mathematical analysis and solver synthesis; AutoNumerics~\citep{du2026autonumerics} combines multi-agent design, implementation, and residual-based verification; and PDE-Controller~\citep{soroco2025pdecontroller} maps natural-language instructions to formal PDE control tasks. These works demonstrate the promise of LLM-driven solver generation, but the solver strategy is still mostly an implicit byproduct of the generated program. AutoPDE instead maintains it as an independently represented object that is built before code generation and revised using numerical evidence.

\section{Preliminaries}
\label{sec:preliminaries}

\subsection{PDEs and the finite element method}
\label{sec:pde-fem}

A PDE relates an unknown field $u$ on a domain $\Omega$ to its derivatives, boundary conditions, and possibly an initial condition. Since most PDEs do not admit closed-form solutions, they must be discretized into finite-dimensional algebraic problems. We focus on the finite element method (FEM), which underlies every case in our benchmark and naturally exposes the strategy choices we study.

As a running example, consider the Helmholtz problem with wave number $k>0$, source $f$, and boundary data $g$:
\begin{equation}\label{eq:helmholtz-strong}
  -\Delta u - k^2 u \;=\; f \quad \text{in } \Omega,
  \qquad
  u \;=\; g \quad \text{on } \partial\Omega .
\end{equation}
FEM converts~\eqref{eq:helmholtz-strong} into executable code in four steps, summarized in Figure~\ref{fig:fem-walkthrough}.

\textbf{(i) Variational formulation.}
Multiplying~\eqref{eq:helmholtz-strong} by a test function $v$ and integrating by parts yields the weak form: find $u \in V$ such that
\begin{equation}\label{eq:helmholtz-weak}
  a(u,v) \;=\; L(v) \quad \forall v \in V,
  \qquad
  a(u,v) = \!\int_\Omega\!\bigl(\nabla u\!\cdot\!\nabla v - k^2 u\,v\bigr)\mathrm{d}x,
  \quad
  L(v) = \!\int_\Omega\! f\,v\,\mathrm{d}x,
\end{equation}
where $V$ is a Sobolev space incorporating the Dirichlet data.

\textbf{(ii) Galerkin discretization.}
The infinite-dimensional space $V$ is replaced by a finite-dimensional subspace $V_h\subset V$ spanned by piecewise polynomial basis functions $\{\phi_i\}_{i=1}^{n}$, e.g.\ continuous Lagrange elements of degree $p$ (``$\mathrm{Lagrange}\,P_p$'') on a triangular mesh of characteristic size $h$. The pair $(h,p)$ controls both accuracy and computational cost.

\textbf{(iii) Algebraic system.}
Writing $u_h = \sum_j u_j \phi_j$ reduces~\eqref{eq:helmholtz-weak} to a sparse linear system
\begin{equation}\label{eq:helmholtz-linsys}
  K\mathbf{u}_h \;=\; \mathbf{b},
  \qquad
  K_{ij} = a(\phi_j,\phi_i),
  \qquad
  b_i = L(\phi_i).
\end{equation}
The \emph{algebraic structure} of $K$ depends on the PDE itself, not on a hyperparameter choice: for the Laplacian alone $K$ is symmetric positive-definite (SPD), but the $-k^2$ reaction term in~\eqref{eq:helmholtz-weak} makes $K$ symmetric \emph{indefinite}.

\textbf{(iv) Algebraic solver.}
The linear system is solved by a Krylov method whose choice is coupled to the algebraic structure: CG requires an SPD operator and diverges for indefinite systems such as Helmholtz~\citep{saad2003iterative}, so a Krylov method compatible with a symmetric indefinite operator (e.g., MINRES, with a symmetric positive-definite preconditioner) is used instead.

These choices are coupled: the weak form sets the achievable approximation, the mesh--degree pair $(h,p)$ controls both error (e.g., $\|u-u_h\|_{L^2}\leq C h^{p+1}|u|_{H^{p+1}}$ for smooth solutions) and system size, and the linear solver turns that system into wall-clock time or convergence failure. Thus a poor strategy choice can inflate error, exceed the time budget, or make the solve fail even if the code itself is syntactically correct. The agent's task is to realize this pipeline in \texttt{solver.py} under the per-case accuracy and runtime budgets described in Section~\ref{sec:exp-setup}.

\begin{figure}[t]
  \centering
  \small
  \begin{minipage}[t]{0.47\linewidth}
    \centering\textbf{Mathematical construction}\\[2pt]
    \rule{\linewidth}{0.4pt}
    \vspace{4pt}

    \colorbox{codebg}{%
      \begin{minipage}{\dimexpr\linewidth-8pt}
        \ttfamily\scriptsize
        \vspace{4pt}

        \textcolor{codecom}{\# (i) variational form}\\
        Find $u\in V$:\ \ $a(u,v)=L(v),\ \forall v\in V$,\\
        $a(u,v)=\int_\Omega(\nabla u\!\cdot\!\nabla v - k^2 uv)\,\mathrm{d}x$,\\
        $L(v)=\int_\Omega f\,v\,\mathrm{d}x.$

        \vspace{6pt}
        \textcolor{codecom}{\# (ii) Galerkin space}\\
        $V_h=\mathrm{span}\{\phi_i\}\subset V$,\ \ $\dim V_h=n$,\\
        Lagrange $P_p$ on a mesh of size $h$.

        \vspace{6pt}
        \textcolor{codecom}{\# (iii) linear system}\\
        $u_h=\sum_j u_j\phi_j\ \Rightarrow\ K\mathbf{u}_h=\mathbf{b}$,\\
        $K_{ij}=a(\phi_j,\phi_i),\ b_i=L(\phi_i)$.

        \vspace{6pt}
        \textcolor{codecom}{\# (iv) algebraic solver}\\
        Krylov $+$ preconditioner matching $K$,\\
        $K$ symmetric \emph{indefinite} (from $-k^2$),\\
        so MINRES (SPD preconditioner), not CG.

        \vspace{4pt}
      \end{minipage}%
    }
  \end{minipage}
  \hfill
  \begin{minipage}[t]{0.50\linewidth}
    \centering\textbf{Executable FEM code}\\[2pt]
    \rule{\linewidth}{0.4pt}
    \vspace{4pt}

\begin{lstlisting}[style=pyfem]
# (i) variational form
u, v = TrialFunction(V), TestFunction(V)
a = (inner(grad(u), grad(v))
     - k*k*inner(u, v)) * dx
L = inner(f, v) * dx

# (ii) Galerkin space
msh = create_square_mesh(N, N)
V   = functionspace(msh, ("Lagrange", p))

# (iii) assemble + apply BC
problem = LinearProblem(a, L, bcs=[bc])

# (iv) solver choice matters
problem.solve(petsc_options={
    "ksp_type": "minres", # NOT "cg"
    "pc_type":  "jacobi", # SPD preconditioner
})
\end{lstlisting}
  \end{minipage}
  \caption{%
    From variational formulation to executable FEM code, illustrated on the Helmholtz equation~\eqref{eq:helmholtz-strong}. Each block on the left corresponds to the code block on its right. The algebraic structure of the discretized system (symmetric indefinite, due to the $-k^2$ reaction term) dictates the choice of linear solver; conjugate gradients, the default for SPD problems, would diverge here. Our agent must make the right choice at every step, for \emph{every} PDE family in the benchmark.%
  }
  \label{fig:fem-walkthrough}
\end{figure}

\section{Motivation}
\label{sec:motivation}

A PDE solver is often treated as a program whose quality can be improved by tuning mesh resolution. This misses the more important choice: before any refinement, the solver commits to a numerical strategy --- discretization, stabilization, linear solver, and resolution policy. For delicate regimes such as advection-dominated transport, saddle-point systems, indefinite operators, or nonlinear couplings, a poor strategy sets an error or runtime ceiling that refinement cannot cheaply overcome.

\paragraph{A concrete example.}
Consider the 2D linear advection equation $u_t+a(x,y)u_x+b(x,y)u_y=0$ on $\Omega=[0,1]^2$ under solid-body rotation, with period $T=1$ so that the exact solution at $t=T$ equals the initial condition. We use the Zalesak slotted disk~\citep{zalesak1979fully}, a standard stress test for convection schemes, and compare first-order donor-cell upwind against fifth-order WENO with a Lax--Friedrichs flux; full setup and mesh sweeps are in Appendix~\ref{app:motivation}.

\begin{figure}[t]
  \centering
  \includegraphics[width=\linewidth]{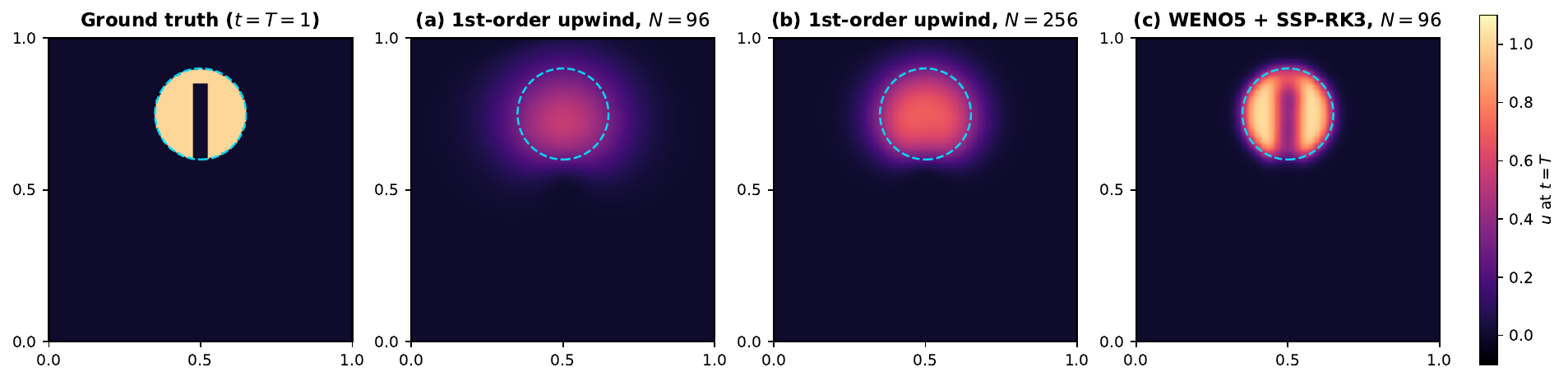}

  \vspace{4pt}
  \small
  \begin{tabular}{@{}llcrr@{}}
    \toprule
    Strategy & Algorithm & $N$ & Time (s) & Rel.\ $L^1$ \\
    \midrule
    (a) 1st-order upwind             & donor-cell upwind + SSP-RK3 & $96$  & $0.60$  & $1.19$ \\
    (b) 1st-order upwind             & donor-cell upwind + SSP-RK3 & $256$ & $10.96$ & $0.95$ \\
    (c) WENO5 + Lax--Friedrichs flux & WENO5 reconstr.\ + SSP-RK3  & $96$  & $5.50$  & $\mathbf{0.46}$ \\
    \bottomrule
  \end{tabular}
  \caption{%
    Ground truth at $t=T=1$ and three explicit numerical strategies on the Zalesak slotted-disk rotation problem; the dashed cyan circle marks where the disk should sit. The table reports the mesh $N$, wall-clock time of the time-stepping loop on a single CPU core, and the relative $L^1$ error against the analytic reference. A full mesh sweep for both algorithms, plotted as (runtime, error) Pareto curves, is given in Figure~\ref{fig:motivation-pareto} (Appendix~\ref{app:motivation}).%
  }
  \label{fig:motivation-strategies}
\end{figure}

\paragraph{Algorithm choice beats mesh refinement.}
Figure~\ref{fig:motivation-strategies} shows that refinement alone does not fix a poor strategy: increasing first-order upwind from $N=96$ to $N=256$ costs roughly $18\times$ more time but only reduces error from $1.19$ to $0.95$, while WENO5 at $N=96$ reaches rel.\ $L^1=0.46$ in $5.50$~s and preserves the slot. The full mesh sweep in Appendix~\ref{app:motivation} confirms the same Pareto ordering. Current code agents struggle with this judgement because failure feedback is routed to code edits, not to the underlying numerical strategy; AutoPDE closes the gap by making that strategy explicit and revisable (Section~\ref{sec:method}).

\section{Method}
\label{sec:method}

\subsection{Overview}

\begin{figure}[t]
  \centering
  \includegraphics[width=\linewidth]{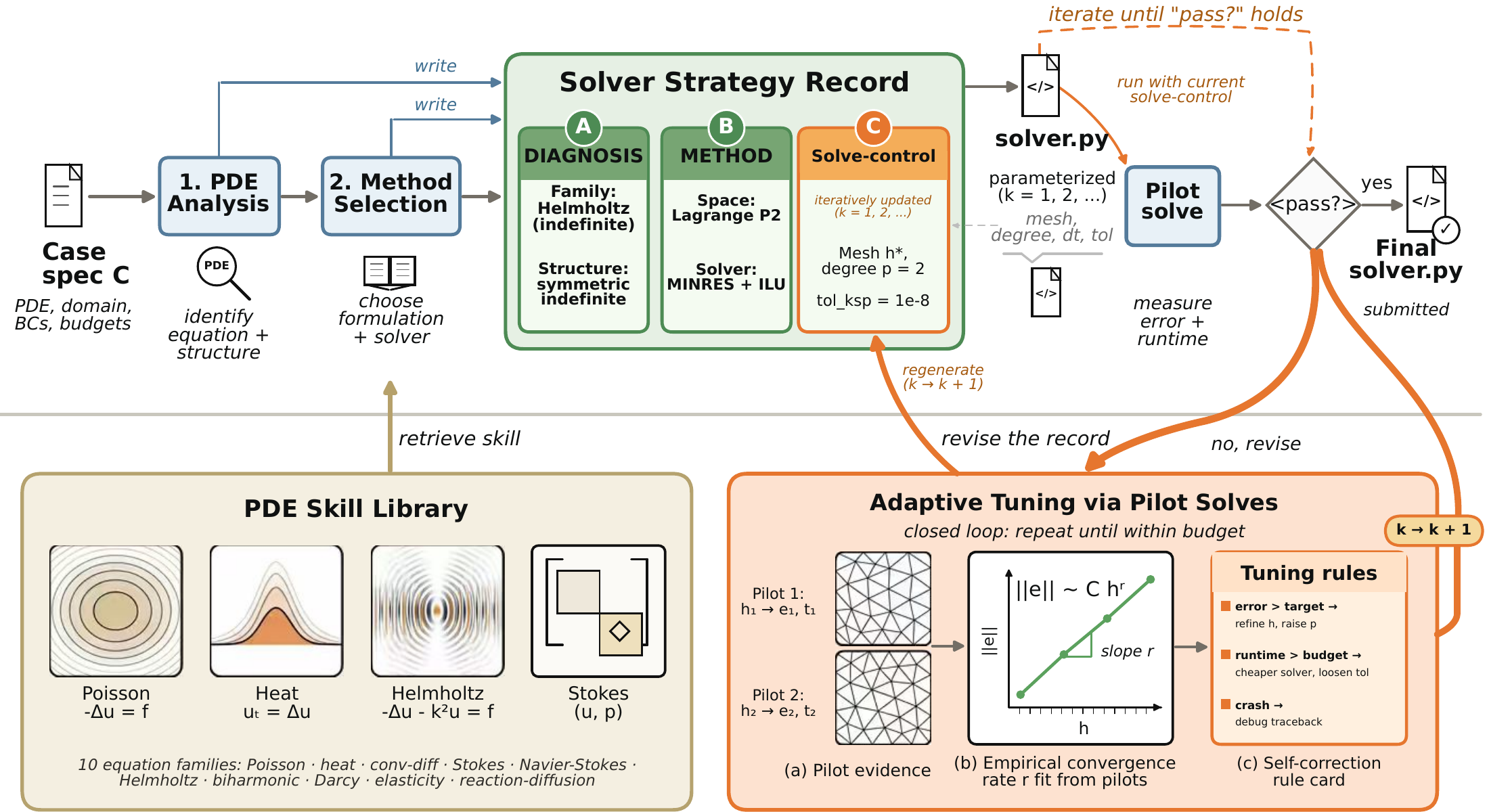}
  \caption{Architecture of AutoPDE. Before code generation, PDE analysis and numerical method selection write the \texttt{DIAGNOSIS} and \texttt{METHOD} cards of the solver-strategy record. After an initial solver is generated, adaptive tuning runs pilot solves and writes measured evidence and updated solve-control choices back to the record.}
  \label{fig:overview}
\end{figure}

We now describe the AutoPDE workflow. The input is a case specification $\mathcal{C}$ describing the PDE, domain, boundary conditions, and evaluation budgets. AutoPDE maintains an explicit solver-strategy record throughout the solve. Before writing code, \emph{PDE analysis} (Section~\ref{sec:pde-analysis}) fills a \texttt{DIAGNOSIS} card, and \emph{numerical method selection} (Section~\ref{sec:skills}) fills a \texttt{METHOD} card. After code generation, \emph{adaptive tuning} (Section~\ref{sec:profiling}) uses pilot runs to refine resolution, element degree, time step, or solver tolerances before final submission. Figure~\ref{fig:overview} shows the architecture.

AutoPDE is implemented on top of \texttt{mini-swe-agent}~\citep{yang2024sweagent}, inheriting its reason--act loop, bash execution, file operations, and runtime API inspection. We add only the strategy-card protocol and two PDE-specific tools: \texttt{get\_pde\_skill}, which retrieves reusable PDE-solving skills, and \texttt{get\_profiling\_guide}, which returns the adaptive testing procedure.

\subsection{PDE Analysis}
\label{sec:pde-analysis}

The system prompt requires the agent to inspect $\mathcal{C}$ and identify the equation family, linearity, time dependence, boundary conditions, dominant mechanism, and expected algebraic structure before writing code. The resulting \texttt{DIAGNOSIS} card distinguishes, for example, convection-dominated transport from diffusion-dominated elliptic problems, and SPD systems from indefinite or saddle-point systems; it also determines which reusable PDE skill is used in the next stage.

\subsection{Numerical Method Selection}
\label{sec:skills}

AutoPDE provides a library of reusable PDE-solving skills covering ten equation families: Poisson, heat, Stokes, Navier--Stokes, Helmholtz, convection--diffusion, biharmonic, Darcy, linear elasticity, and reaction--diffusion. Each skill is a structured plain-text guide, not executable code, and describes typical formulations, discretizations, stabilization choices, boundary treatments, solvers, preconditioners, and pitfalls.

Using the \texttt{DIAGNOSIS} card, the agent calls \texttt{get\_pde\_skill} for the identified family and writes a \texttt{METHOD} card committing to a formulation, finite-element space, stabilization, algebraic solver, and preconditioner. Typical entries include Taylor--Hood $P_2$--$P_1$ elements for Stokes, SUPG for advection-dominated convection--diffusion, or a non-CG Krylov method for indefinite Helmholtz systems.

\subsection{Adaptive Tuning}
\label{sec:profiling}

After the initial solver is generated, AutoPDE uses a rule-based profiling guide rather than a learned or closed-form extrapolator. The agent is instructed to run the solver, print the measured error and wall time when available, compare them against the per-case targets, and self-correct along fixed directions: refine the mesh or raise polynomial degree when error is too high, simplify the solve or adjust tolerances when runtime is too high, and repair tracebacks when execution fails. The solver is submitted only after the latest run satisfies the target metrics or no further local repair is possible.

\section{Experiments}

\subsection{Setup}
\label{sec:exp-setup}

\paragraph{Benchmark.}
We evaluate AutoPDE on \emph{PDE Agent Bench}~\citep{huang2026pdeagentbench}, a benchmark suite for PDE solver generation that covers eight equation families spanning the main regimes of classical PDE solving: heat, convection--diffusion, Stokes, Navier--Stokes, Helmholtz, biharmonic, linear elasticity, and reaction--diffusion, for a total of $191$ cases. Each case specifies a PDE together with the accuracy and time budget under which it is to be solved, and is evaluated by executing the agent's generated solver in an isolated environment and comparing against a per-case reference. Case counts, construction of references, and the exact scoring pipeline are detailed in Appendix~\ref{app:benchmark}; results on an additional Poisson supplement are reported separately in Appendix~\ref{app:poisson}.

\paragraph{Baselines.}
We compare AutoPDE against three classes of baselines.
\emph{(i)~Naive LLM.}
Single-shot generation from the same benchmark prompt with no tool use or execution feedback. We sweep four backbones, DeepSeek~V3.2, Gemini~2.5~Pro, GPT~5.1, and Claude~Opus~4.6, to separate the contribution of backbone capability from that of agent scaffolding.
\emph{(ii)~General-purpose coding agents.}
\texttt{mini-swe-agent}~\citep{yang2024sweagent}, a minimal agentic scaffold built around a bash-execution loop with no PDE-specific knowledge, and \texttt{OpenHands}~\citep{wang2024openhands}, a more feature-rich code agent. Both expose only generic file/shell tools.
\emph{(iii)~PDE-specific agent.}
\texttt{CodePDE}~\citep{li2025codepde}, a PDE-oriented inference framework that originally combines code generation, self-debugging, refinement, and test-time scaling. We re-target it to our evaluation harness; its exact configuration is described in Appendix~\ref{app:baselines}.
To isolate scaffolding from backbone, all three agent baselines (\texttt{mini-swe-agent}, \texttt{OpenHands}, \texttt{CodePDE}) and AutoPDE are run on \emph{both} Claude~Opus~4.6 and GPT~5.1, so that any gap between scaffolds at fixed backbone reflects the scaffolding rather than the language model. Full baseline configurations and prompt templates are in Appendix~\ref{app:baselines}. Ablations of AutoPDE's own components are reported separately in Section~\ref{sec:ablation}.

\paragraph{Metrics.}
The primary metric is the end-to-end task pass rate. A case is counted as passed only if the generated solver (1)~executes successfully, (2)~its relative $L^2$ error against the reference is below $\mathrm{target\_error}$, and (3)~its runtime stays below $\mathrm{target\_time}$. Let $e^\star$ and $t^\star$ be the oracle error and runtime, and let $\tau_\mathrm{acc}, \tau_\mathrm{time}$ be the per-case tolerance multipliers shipped with the benchmark. The targets are
\[
  \mathrm{target\_error} = \max\!\left(e^\star \cdot \tau_\mathrm{acc},\; 10^{-6}\right), \qquad \mathrm{target\_time} = t^\star \cdot \tau_\mathrm{time}.
\]

\subsection{Main Results}

Table~\ref{tab:main-results} summarizes the main comparison against baselines across all eight equation families of PDE Agent Bench.

\begin{table}[t]
  \centering
  \caption{%
    Main pass-rate results on PDE Agent Bench ($191$ cases across eight equation families). Each cell is the percentage of passed cases within that family and the \emph{Overall} column aggregates over all $191$. Column headers use short codes for brevity: Heat (heat), CD (convection--diffusion), Stokes, NS (Navier--Stokes), Helm (Helmholtz), BH (biharmonic), LE (linear elasticity), RD (reaction--diffusion). At fixed backbone, gaps between AutoPDE and the other agents reflect scaffolding rather than model capability. \textbf{Bold} marks the overall best in each column, \underline{underline} marks the second-best (ties are all bolded).%
  }
  \label{tab:main-results}
  \small
  \setlength{\tabcolsep}{4pt}
  \begin{tabular}{@{}llcccccccc|c@{}}
    \toprule
    Backbone & Method & Heat & CD & Stokes & NS & Helm & BH & LE & RD & Overall \\
    \midrule
    \multirow{4}{*}{Claude~Opus~4.6}
        & mini-swe-agent      & 30.0 &  0.0 &  0.0 &  0.0 &  5.3 & \underline{93.3} &  0.0 & \underline{68.2} & 22.0 \\
        & OpenHands           & 37.5 &  0.0 &  5.6 &  3.6 &  0.0 & \textbf{100.0} & 42.9 & 63.6 & 25.7 \\
        & CodePDE             &  0.0 &  9.5 & 11.1 & \textbf{50.0} & \textbf{94.7} & 80.0 & \underline{85.7} & 22.7 & 31.9 \\
        & \textbf{AutoPDE}    & 40.0 & 33.3 & \underline{44.4} & \underline{46.4} & \underline{89.5} & \textbf{100.0} & 71.4 & \textbf{72.7} & \textbf{54.5} \\
    \midrule
    \multirow{4}{*}{GPT~5.1}
        & mini-swe-agent      & 37.5 & 38.1 &  0.0 & 35.7 & 84.2 & 26.7 & \underline{85.7} & 27.3 & 38.2 \\
        & OpenHands           & \underline{45.0} & 50.0 &  0.0 & 14.3 & 68.4 & 53.3 & 71.4 & 36.4 & \underline{40.3} \\
        & CodePDE             &  5.0 & 14.3 &  0.0 &  0.0 & 36.8 & 13.3 & 42.9 &  4.5 & 11.0 \\
        & \textbf{AutoPDE}    & \textbf{60.0} & \textbf{59.5} & 22.2 & 35.7 & 78.9 & 53.3 & \textbf{100.0} & 50.0 & \textbf{54.5} \\
    \midrule
    DeepSeek V3.2   & Naive LLM               & 12.5 &  2.4 &  5.6 &  0.0 &  0.0 & 20.0 & 42.9 &  0.0 &  6.8 \\
    Gemini 2.5 Pro  & Naive LLM               & 20.0 &  2.4 &  0.0 &  0.0 &  0.0 & 33.3 & 14.3 &  4.5 &  8.4 \\
    GPT 5.1         & Naive LLM               & 20.0 & 23.8 &  0.0 &  0.0 & 68.4 & 33.3 & 71.4 &  0.0 & 21.5 \\
    Claude Opus 4.6 & Naive LLM               &  0.0 &  0.0 & \textbf{66.7} & 42.9 & 26.3 & \textbf{100.0} & 42.9 &  4.5 & 25.1 \\
    \bottomrule
  \end{tabular}
\end{table}

\paragraph{Overall pass rate.}
AutoPDE reaches an overall pass rate of $54.5\%$ on \emph{both} Claude~Opus~4.6 and GPT~5.1, tied for the best result in Table~\ref{tab:main-results}. The strongest non-AutoPDE baseline is \texttt{OpenHands} on GPT~5.1 at $40.3\%$, so AutoPDE leads by $14.2$ percentage points; against the best Claude-backbone baseline (\texttt{CodePDE} at $31.9\%$) the gap grows to $22.6$ percentage points. Two finer patterns are worth noting. First, AutoPDE \emph{flattens} the dependence on the backbone: the gap between its two backbones is $0.0$ percentage points, while all three other agent families show substantial gaps of $15$--$21$ percentage points depending on the backbone (e.g., \texttt{CodePDE} drops from $31.9\%$ on Claude to $11.0\%$ on GPT, and within the Naive LLM group the range is $6.8$--$25.1\%$). The interpretation is that explicit, structured solver-strategy reasoning lets a weaker code-writing backbone reach the same end-to-end quality as a stronger one. Second, on the same backbone the four agent scaffolds (\texttt{mini-swe-agent}, \texttt{OpenHands}, \texttt{CodePDE}, AutoPDE) are exposed to the same model and the same per-case prompt; the spread between them ($22.0$--$54.5\%$ on Claude, $11.0$--$54.5\%$ on GPT) is therefore attributable to scaffolding rather than to model capability.

\paragraph{Per-family trends.}
The most informative reading of Table~\ref{tab:main-results} is that AutoPDE is the only method whose pass rate stays non-trivial in every family: its minimum is $22.2\%$ (Stokes, GPT backbone), whereas every other configuration has at least one family with a pass rate at or near zero. The contrast is sharpest on families where success requires a non-trivial \emph{numerical} decision rather than a textbook recipe. Heat and convection--diffusion both fall in this category: AutoPDE on GPT~5.1 tops them at $60.0\%$ and $59.5\%$, while all baselines on Claude score $\leq 37.5\%$ and $\leq 9.5\%$ respectively, and even a strong single-shot Claude baseline collapses to $0.0\%$ on both. On Stokes and Helmholtz the leaderboard is split: Naive Claude (which happens to default to a textbook mixed formulation) tops Stokes at $66.7\%$ with AutoPDE second at $44.4\%$, and CodePDE tops Helmholtz at $94.7\%$ with AutoPDE second at $89.5\%$. We read this as evidence that \emph{some} families are effectively solved by one standard recipe that a single scaffold happens to embed; AutoPDE closes in on those specialists without dominating them. Finally on the more boilerplate-friendly families (biharmonic, linear elasticity, reaction--diffusion), AutoPDE either matches or leads ($100\%$ biharmonic tied with OpenHands~Claude and Naive~Claude, $100\%$ linear elasticity on GPT, $72.7\%$ reaction--diffusion), confirming that the regimes where AutoPDE shines most are exactly those where \emph{implicit} code-first agents commit strategy-level mistakes.

\subsection{Analysis}
\label{sec:analysis}

To understand \emph{why} AutoPDE out-performs generic code agents on the benchmark's hardest family, we use the 41 convection--diffusion cases, which span four orders of magnitude in P\'eclet number ($\mathrm{Pe}=\max|\boldsymbol{\beta}|/\varepsilon$). All five Claude-backbone runs are analysed on the same cases; we read their generated \texttt{solver.py} and bucket each case by $\mathrm{Pe}$, then measure (a) whether any streamline-upwind / SUPG stabilisation token appears in the form, and (b) whether the case ultimately passes the benchmark oracle.

\begin{figure}[t]
  \centering
  \includegraphics[width=0.95\linewidth]{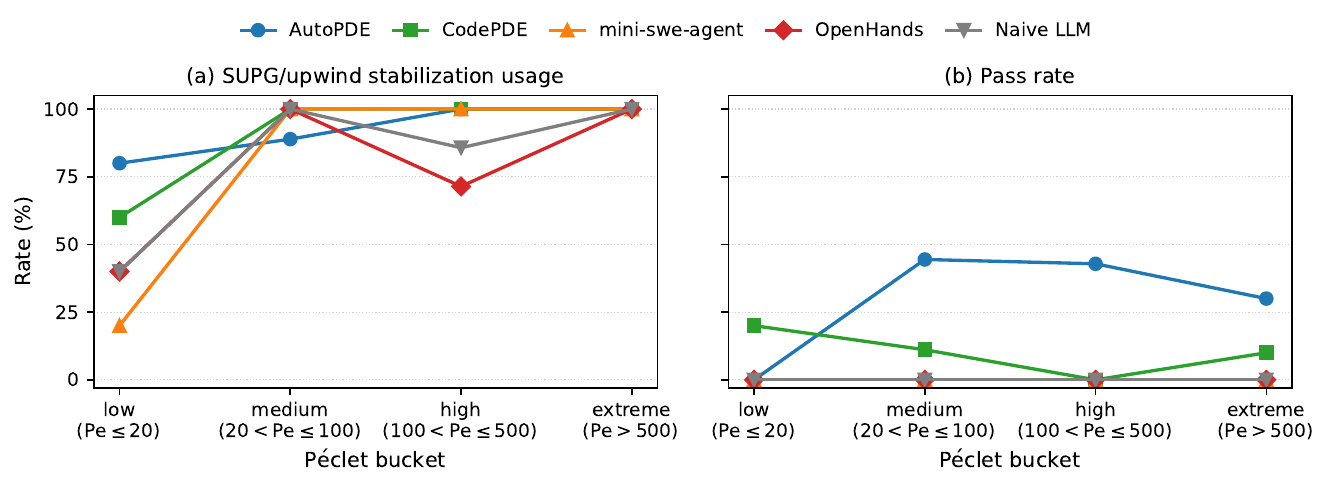}
  \caption{P\'eclet-stratified behaviour of five Claude-backbone methods on the 41 convection--diffusion cases. \textbf{(a)} Fraction of generated solvers that contain a streamline-upwind / SUPG stabilisation term. \textbf{(b)} Fraction of cases that pass the oracle. All methods almost always \emph{write} stabilisation once $\mathrm{Pe}>20$, but only AutoPDE converts that into non-trivial pass rate at every $\mathrm{Pe}$ level.}
  \label{fig:pe-stab}
\end{figure}

\paragraph{Stabilisation usage says little; the joint discretisation choice decides.}
Figure~\ref{fig:pe-stab}(a) shows that by $\mathrm{Pe}>20$ every method already reaches $\approx 100\%$ SUPG-usage and stays there, so the baselines are \emph{not} failing because they forgot to stabilise. Yet in panel~(b) AutoPDE is the only method with a non-trivial pass rate across convection-dominated buckets ($0/44/43/30\%$ on low/medium/high/extreme), while mini-swe-agent, OpenHands, and Naive LLM remain at $0\%$ pass everywhere despite writing SUPG on $80$--$100\%$ of cases. What makes convection--diffusion a stress test is that the relevant knobs are mutually constrained: the boundary-layer width $\delta\sim\varepsilon/|\boldsymbol\beta|$ must be resolved by the mesh, the cell-P\'eclet number $\mathrm{Pe}_c=|\boldsymbol\beta|h/(2\varepsilon)$ determines whether the Hughes optimal stabilisation parameter $\tau=(h/(2|\boldsymbol\beta|))(\coth\mathrm{Pe}_c - 1/\mathrm{Pe}_c)$ reduces to its convection-dominated limit $h/(2|\boldsymbol\beta|)$ or not, and polynomial degree $p$ shifts the effective $h_{\mathrm{eff}}=h/p$ that appears in both conditions. Picking $\tau$ without knowing $(h,p)$, or choosing $(h,p)$ without knowing $\mathrm{Pe}$, is genuinely ill-posed; a method can "write SUPG'' and still produce a solver in which the three knobs contradict each other. This is precisely the coupled decision AutoPDE's \emph{PDE Analysis}$\to$\emph{Numerical Method Selection} stages are designed to surface before any code is emitted.

\paragraph{Evidence from the generated solvers.}
Inspecting the $41\times 5=205$ \texttt{solver.py} files confirms that the pass-rate gap is not explained by any single knob. None of the five methods uses an SPD solver such as CG on the non-symmetric convection--diffusion system; all select GMRES, BiCGSTAB, or direct LU, so linear algebra is not where they split. Most methods also escalate mesh resolution with~$\mathrm{Pe}$: the median $N$ rises from low to extreme $\mathrm{Pe}$ for AutoPDE ($64\!\to\!100$), CodePDE ($64\!\to\!128$), mini-swe-agent ($85\!\to\!128$), OpenHands ($56\!\to\!100$), and Naive LLM ($80\!\to\!128$). The qualitative discriminator is \emph{range}: AutoPDE's mesh spans $N\in[48,500]$ and FE degree $\{1,2,3\}$, whereas OpenHands stays within $N\in[32,128]$, CodePDE within $N\in[64,200]$, and mini-swe / Naive within $N\in[64,200]$. AutoPDE is thus the only method that, for a non-trivial sub-population, commits to the resolution the cell-P\'eclet condition requires --- the cases in panel~(b) that pass. Since accuracy-stage failures are near zero once a solver reaches the oracle, the gap reflects whether the generated $(\tau,h,p)$ triple is internally consistent with~$\mathrm{Pe}$, or contradicts itself and aborts.

\subsection{Ablation Studies}
\label{sec:ablation}

We ablate the three methodological stages of AutoPDE on a focused subset of four PDE families --- convection--diffusion (CD, $42$ cases), Stokes ($18$), Navier--Stokes (NS, $28$), and Helmholtz (Helm, $19$), for a total of $107$ cases --- chosen because they are the families where the full AutoPDE gain over a generic code agent is largest in Table~\ref{tab:main-results}, so drops are easy to read. All variants use the same Claude~Opus~4.6 backbone as the "AutoPDE (full)'' row. Starting from full AutoPDE (variant~A) we independently disable each stage: (B)~\emph{w/o DIAGNOSIS}, removing the PDE-analysis card; (C)~\emph{w/o PDE skills}, removing the method-selection card together with the skill library; (D)~\emph{w/o Adaptive Tuning}, disabling the profiling-driven pilot-solve refinement.

\begin{table}[t]
  \centering
  \caption{%
    Ablation of AutoPDE's three methodological stages on a $107$-case subset spanning four families (Claude~Opus~4.6 backbone). Per-family cells report pass rate in percent; the rightmost block reports pass and exec-valid rates on the full $107$-case subset. Row A is the full AutoPDE configuration, reproduced from Table~\ref{tab:main-results}.%
  }
  \label{tab:ablation}
  \small
  \setlength{\tabcolsep}{4pt}
  \begin{tabular}{@{}lcccc|cc@{}}
    \toprule
    \multirow{2}{*}{Variant} & \multicolumn{4}{c|}{Pass rate by family (\%)} & \multicolumn{2}{c}{Overall ($107$ cases)} \\
     & CD & Stokes & NS & Helm & Pass (\%) & Exec (\%) \\
    \midrule
    \textbf{A. AutoPDE (full)}       & \textbf{33.3} & \textbf{44.4} & \textbf{46.4} & \textbf{89.5} & \textbf{48.6} & \textbf{57.9} \\
    B. w/o DIAGNOSIS (S1)            & 23.8 & 38.9 & 35.7 & 78.9 & 39.3 & 47.7 \\
    C. w/o PDE skills (S2)           &  9.5 &  5.6 & 14.3 & 31.6 & 14.0 & 21.5 \\
    D. w/o Adaptive Tuning (S3)      & 23.8 & 33.3 & 28.6 & 84.2 & 37.4 & 57.0 \\
    \bottomrule
  \end{tabular}
\end{table}

Every stage contributes positively: removing any one lowers the overall pass rate relative to the full configuration (A, $48.6\%$) across all four families. The largest drop comes from S2 (PDE skills, $34.6$ percentage points), followed by S3 (Adaptive Tuning, $11.2$) and S1 (DIAGNOSIS, $9.3$). S3 leaves the exec rate almost unchanged ($57.9\%\!\to\!57.0\%$) while still costing $11.2$ percentage points of pass rate, consistent with its role of refining an already-executable solver rather than producing one.

\section{Conclusion}

We presented AutoPDE, a PDE solving agent that makes the solver strategy explicit before generating code, using PDE analysis, numerical method selection, and adaptive tuning built from reusable PDE skills. On PDE Agent Bench, AutoPDE reaches a pass rate of $54.5\%$, improving over the strongest baseline by $14.2$ percentage points.

\newpage
\small
\bibliographystyle{plainnat}
\bibliography{references}

\newpage
\normalsize
\appendix

\section{Limitations}

Our current benchmark and skill library are built on top of the FEniCSx / dolfinx stack, and all experiments in this paper evaluate agents through that single backend. Extending AutoPDE to other computational-mathematics libraries (e.g.\ Firedrake, deal.II, and classical finite-difference or spectral stacks) is a natural direction for future work.

\section{Broader Impact}

AutoPDE targets the automated generation of numerical PDE solvers, a task that lies within scientific and engineering computing. By making the underlying solver strategy explicit and inspectable, the system helps lower the entry barrier for non-specialists to obtain reasonable numerical solutions to classical PDE problems, and can accelerate prototyping workflows in research and education. We do not foresee direct negative societal impacts beyond those common to code-generating systems: generated solvers may occasionally be numerically unreliable on problems outside the tested families, and users should validate outputs against trusted references before relying on them in safety-critical settings. All experiments in this paper use synthetic PDE benchmarks with no personal or sensitive data.

\section{Benchmark Details}
\label{app:benchmark}

\paragraph{Case format.}
Each case in PDE Agent Bench is a JSON object specifying \texttt{oracle\_config} (PDE type, coefficients, domain, mesh, FE family/degree, boundary conditions, output grid, solver options) and \texttt{evaluation\_config} (timeout, accuracy tolerance, time tolerance). For manufactured-solution cases, the prompt includes the symbolic exact solution; for no-exact-solution cases, the oracle pipeline constructs a higher-fidelity reference.

\paragraph{Oracle construction.}
The oracle solver dispatches to a PDE-family-specific routine (Poisson, heat, Stokes, etc.), solves on the reference mesh, and caches the result as \texttt{.json} + \texttt{.npz}. The evaluator then interpolates both agent and oracle solutions onto a common uniform grid before computing relative $L^2$ error.

\paragraph{Case distribution.}
The $191$ cases of PDE Agent Bench break down as: heat (40), convection--diffusion (42), Stokes (18), Navier--Stokes (28), Helmholtz (19), biharmonic (15), linear elasticity (7), reaction--diffusion (22). An additional $50$ Poisson cases are reported separately as a supplementary experiment in Appendix~\ref{app:poisson}.

\paragraph{Per-family results.}
Table~\ref{tab:per-family-exec} reports the per-family exec-valid rate on the $191$-case benchmark, complementing the pass-rate numbers in Table~\ref{tab:main-results} of the main paper. Exec-valid rate counts any case whose generated solver runs to completion, regardless of whether its accuracy or runtime gates pass.

\begin{table}[t]
  \centering
  \caption{Per-family exec-valid rate (\%) on the $191$-case benchmark, corresponding to the pass-rate Table~\ref{tab:main-results}. Column codes: Heat, CD (convection--diffusion), Stokes, NS (Navier--Stokes), Helm (Helmholtz), BH (biharmonic), LE (linear elasticity), RD (reaction--diffusion).}
  \label{tab:per-family-exec}
  \scriptsize
  \setlength{\tabcolsep}{5pt}
  \begin{tabular}{@{}llcccccccc|c@{}}
    \toprule
    Backbone & Method & Heat & CD & Stokes & NS & Helm & BH & LE & RD & Overall \\
    \midrule
    \multirow{4}{*}{Claude~Opus~4.6}
        & mini-swe-agent      & 30.0 &  0.0 &  0.0 &  3.6 &  5.3 & 100.0 &  0.0 & 100.0 & 26.7 \\
        & OpenHands           & 37.5 &  0.0 &  5.6 &  3.6 &  0.0 & 100.0 & 42.9 & 95.5 & 29.3 \\
        & CodePDE             &  0.0 &  9.5 & 55.6 & 57.1 & 100.0 & 86.7 & 85.7 & 50.0 & 41.4 \\
        & \textbf{AutoPDE}    & 40.0 & 33.3 & 77.8 & 60.7 & 89.5 & 100.0 & 85.7 & 90.9 & 62.3 \\
    \midrule
    \multirow{4}{*}{GPT~5.1}
        & mini-swe-agent      & 50.0 & 66.7 & 16.7 & 64.3 & 89.5 & 66.7 & 85.7 & 72.7 & 61.8 \\
        & OpenHands           & 52.5 & 69.0 &  0.0 & 14.3 & 68.4 & 60.0 & 71.4 & 50.0 & 48.2 \\
        & CodePDE             &  5.0 & 21.4 &  0.0 &  0.0 & 36.8 & 40.0 & 42.9 &  9.1 & 15.2 \\
        & \textbf{AutoPDE}    & 65.0 & 73.8 & 22.2 & 39.3 & 89.5 & 80.0 & 100.0 & 68.2 & 64.4 \\
    \midrule
    DeepSeek V3.2   & Naive LLM & 30.0 &  2.4 & 11.1 &  0.0 &  0.0 & 33.3 & 57.1 &  9.1 & 13.6 \\
    Gemini 2.5 Pro  & Naive LLM & 20.0 &  4.8 &  0.0 &  0.0 &  5.3 & 40.0 & 14.3 &  4.5 &  9.9 \\
    GPT 5.1         & Naive LLM & 32.5 & 33.3 &  0.0 &  0.0 & 73.7 & 60.0 & 71.4 &  9.1 & 29.8 \\
    Claude Opus 4.6 & Naive LLM &  0.0 &  0.0 & 88.9 & 53.6 & 26.3 & 100.0 & 42.9 & 22.7 & 30.9 \\
    \bottomrule
  \end{tabular}
\end{table}

\section{Poisson Supplementary Results}
\label{app:poisson}

Because the Poisson equation is well conditioned and admits a textbook CG+AMG recipe, every scaffold we evaluate already reaches a high pass rate on it; including Poisson in the main table would therefore dilute rather than inform the scaffolding comparison. For completeness we report Poisson results separately in Table~\ref{tab:poisson-supplement} as a sanity-check supplementary experiment: $50$ Poisson cases solved under the same prompt, evaluator, and per-case wall-clock budget as the main benchmark. Every Claude-backbone agent (\texttt{mini-swe-agent}, \texttt{OpenHands}, \texttt{CodePDE}, AutoPDE) and the single-shot Claude~Opus~4.6 call pass $\geq 84\%$ of the Poisson cases, confirming that the Poisson column does not discriminate between scaffolds.

\begin{table}[t]
  \centering
  \caption{Supplementary Poisson results ($50$ cases). \emph{Pass} is the fraction satisfying accuracy and time gates; \emph{Exec} is the fraction whose generated solver runs to completion. All agent scaffolds on Claude reach $\geq 84\%$ pass rate, so Poisson is excluded from the main comparison in Table~\ref{tab:main-results}.}
  \label{tab:poisson-supplement}
  \small
  \setlength{\tabcolsep}{6pt}
  \begin{tabular}{@{}llcc@{}}
    \toprule
    Backbone & Method & Pass (\%) & Exec (\%) \\
    \midrule
    \multirow{4}{*}{Claude~Opus~4.6}
        & mini-swe-agent      & 98.0 & 98.0 \\
        & OpenHands           & 94.0 & 94.0 \\
        & CodePDE             & 92.0 & 92.0 \\
        & \textbf{AutoPDE}    & 84.0 & 84.0 \\
    \midrule
    \multirow{4}{*}{GPT~5.1}
        & mini-swe-agent      & 78.0 & 82.0 \\
        & OpenHands           & 62.0 & 62.0 \\
        & CodePDE             & 32.0 & 38.0 \\
        & \textbf{AutoPDE}    & 82.0 & 82.0 \\
    \midrule
    DeepSeek V3.2   & Naive LLM & 38.0 & 54.0 \\
    Gemini 2.5 Pro  & Naive LLM & 48.0 & 52.0 \\
    GPT 5.1         & Naive LLM & 78.0 & 82.0 \\
    Claude Opus 4.6 & Naive LLM & 100.0 & 100.0 \\
    \bottomrule
  \end{tabular}
\end{table}

\section{Motivation Benchmark Details}
\label{app:motivation}

This appendix expands the motivation example of Section~\ref{sec:motivation}.

\paragraph{Problem.}
We solve the two-dimensional linear advection equation under a solid-body rotation,
\begin{equation*}
  u_t + a(x,y)\,u_x + b(x,y)\,u_y = 0 \quad \text{in } \Omega=[0,1]^2, \qquad a(x,y) = -2\pi(y-0.5), \quad b(x,y) = +2\pi(x-0.5),
\end{equation*}
for $t \in [0, T]$ with $T=1$. Zero-Dirichlet boundary conditions are imposed on $\partial\Omega$; they are consistent because the initial profile has compact support strictly inside $\Omega$ and the velocity field is tangent to the unit square. The rotation has period $T=1$, so the analytic solution at $t=T$ coincides with the initial condition. The initial condition is the Zalesak slotted disk~\citep{zalesak1979fully}: the indicator of the disk of radius $r=0.15$ centered at $(0.5, 0.75)$ with a vertical slot of half-width $0.025$ and top offset $0.10$ removed.

\paragraph{Algorithms.}
All three strategies use a cell-centred uniform grid on $\Omega$ with $N \times N$ cells, the Shu--Osher SSP-RK3 time integrator, and a CFL number of $0.4$ (with $\Delta t$ set from $\max\sqrt{a^2+b^2}$). They differ only in the convection discretization:
\begin{itemize}[leftmargin=1.5em]
  \item Strategies (a) and (b) use first-order donor-cell upwind differencing, i.e.\ $u_x \approx (u_{i,j} - u_{i-1,j})/h$ when $a_{i,j} > 0$ and the analogous forward difference otherwise, and likewise for $u_y$.
  \item Strategy (c) uses a fifth-order WENO reconstruction~\citep{liu1994weighted,jiang1996efficient} of the interface values with a Lax--Friedrichs numerical flux.
\end{itemize}

\paragraph{Error metric.}
Since the analytic solution at $t=T$ is again the initial slotted disk, we report the relative $L^1$ error on a fixed $200\times200$ evaluation grid,
\(
  \mathrm{rel}\text{-}L^1 = \sum_{i,j} | u^{\mathrm{num}}_{i,j} - u^{\star}_{i,j} | \,/\, \sum_{i,j} | u^{\star}_{i,j} |,
\)
where $u^{\star}$ is the analytic slotted disk. Because the support of $u^{\star}$ is a small fraction of $\Omega$, numerical diffusion easily inflates $\mathrm{rel}\text{-}L^1$ above $1$; the absolute magnitude of the error is therefore less informative than the ordering, which mirrors what is visible in Figure~\ref{fig:motivation-strategies}.

\paragraph{Wall-clock time.}
All wall-clock numbers in Figure~\ref{fig:motivation-strategies} are measured on a single CPU core and cover the time-stepping loop only (SSP-RK3 sub-stages plus the RHS evaluation); initial-condition construction and the final interpolation to the evaluation grid are excluded. For the Pareto plot we sweep the mesh resolution $N \in \{32, 64, 96, 128, 160, 192, 256, 320, 384\}$ for the first-order upwind algorithm and $N \in \{32, 48, 64, 96, 128\}$ for WENO5, collecting a single (runtime, error) point per configuration.

\begin{figure}[t]
  \centering
  \includegraphics[width=0.55\linewidth]{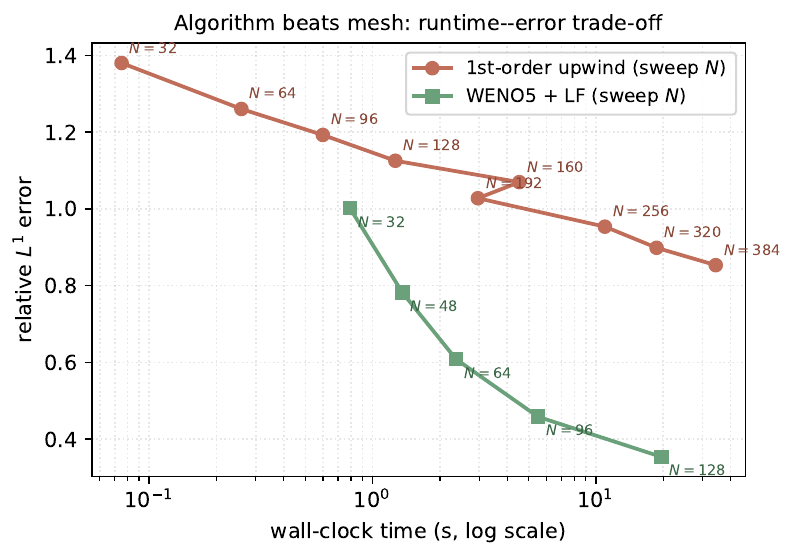}
  \caption{(Runtime, rel.\ $L^1$) Pareto curves for the two numerical strategies of Section~\ref{sec:motivation}, obtained by sweeping the mesh $N$ as listed above. The WENO5 curve lies strictly below the first-order upwind curve at every runtime budget we tried: WENO5 at $N=48$ already reaches rel.\ $L^1 \approx 0.78$ in $1.36$~s, while the finest upwind run we measured, $N=384$, needs $34.4$~s to reach only $0.85$ ($25\times$ the wall-clock time for a worse solution). The gap does not close under further refinement because first-order upwind carries an $O(h)$ numerical diffusion that caps its achievable error for any given runtime budget.}
  \label{fig:motivation-pareto}
\end{figure}

\section{Baselines Details}
\label{app:baselines}

This appendix documents the exact configuration of each baseline. Every baseline uses the same per-case prompt supplied by PDE Agent Bench (PDE specification, interface contract, and pass criteria), and writes its final output into a per-case \texttt{solver.py}. Per-case wall-clock timeout is $600$~s throughout. All agent baselines (\texttt{mini-swe-agent}, \texttt{OpenHands}, \texttt{CodePDE}, AutoPDE) are evaluated on \emph{both} Claude~Opus~4.6 and GPT~5.1; the Naive LLM configuration additionally sweeps two more backbones, DeepSeek~V3.2 and Gemini~2.5~Pro.

\paragraph{Naive LLM.}
Single-shot code generation. The prompt is sent once to the backbone via a chat-completion call; the model's reply is parsed for a fenced Python block, written verbatim to \texttt{solver.py}, and executed. No tools, no feedback loop, no retries. Backbones evaluated: DeepSeek~V3.2, Gemini~2.5~Pro, GPT~5.1, and Claude~Opus~4.6. All runs use temperature $0.0$ and \texttt{max\_tokens}$=32{,}768$.

\paragraph{mini-swe-agent.}
We use the official \texttt{mini-swe-agent} package (\texttt{minisweagent.agents.interactive.InteractiveAgent}) with its built-in \texttt{mini.yaml} configuration, switched to fully autonomous (\texttt{yolo}) mode so that no human confirmation is needed between actions. For each case we create a fresh temporary workspace, instruct the agent to write the final solution to \texttt{./solver.py}, and hand it the benchmark prompt as the task description. We use temperature $0.2$, \texttt{max\_tokens}$=32{,}768$, and the default mini-swe-agent per-case budgets of $50$ steps and \$3 of model cost, which are the dominant limits in practice. We keep the substrate's default tool surface, which already includes bash execution and the \texttt{inspect} runtime API introspection tool; no PDE-specific tools are added on top.

\paragraph{OpenHands.}
We use the OpenHands code-agent platform~\citep{wang2024openhands} through its public Python API, configured for fully autonomous execution (no human-in-the-loop confirmations). Each case is run in a fresh sandbox with the standard OpenHands tool surface (bash, file editing, code execution); the agent is instructed to produce a single \texttt{solver.py} matching the benchmark interface contract, and to confirm completion via a structured signal. We use temperature $0.2$ and the platform's default per-case step and cost budgets; configurations otherwise mirror those of \texttt{mini-swe-agent} so that the difference between the two reflects the substrate, not the prompt.

\paragraph{CodePDE.}
We re-target the CodePDE pipeline~\citep{li2025codepde} to the same per-case prompt interface. CodePDE was originally proposed as a composite inference framework combining (a)~code generation from a PDE-aware system prompt augmented with few-shot solver templates, (b)~self-debugging driven by execution feedback, (c)~seed-based iterative refinement, and (d)~test-time scaling through best-of-$n$ sampling. In our runs we enable (a) together with a pared-down version of (d): each case is generated with $N=3$ repeated samples using a PDE-family-specific few-shot DOLFINx template as context, and the final \texttt{solver.py} is picked from those $3$ candidates; debugging (b) and refinement (c) are not enabled, so that CodePDE's total number of LLM calls per case is comparable to the other baselines (which make a single autonomous call each). Sampling uses temperature $0.2$ and \texttt{max\_tokens}$=32{,}768$. We use the CodePDE prompts and templates as released, with only the output-contract string (module-level \texttt{solve(case\_spec) -> dict}) aligned to the benchmark interface so that the evaluator can invoke the produced solver.

\paragraph{AutoPDE.}
AutoPDE is built on the same \texttt{mini-swe-agent} substrate as the \texttt{mini-swe-agent} baseline above, and is run through the \texttt{ustcpdeagent} wrapper (see also Appendix~\ref{app:impl}). Inherited from the substrate are the bash-execution loop and the \texttt{inspect} runtime API introspection tool. On top of this, AutoPDE adds two PDE-specific tools, \texttt{get\_pde\_skill} (retrieval from the PDE skill library) and \texttt{get\_profiling\_guide} (the adaptive-tuning protocol of Section~\ref{sec:profiling}), together with a system-prompt scaffolding that elicits structured \texttt{DIAGNOSIS} and \texttt{METHOD} cards before any code is written. Because AutoPDE and the \texttt{mini-swe-agent} baseline share the substrate, any performance gap between them isolates the contribution of these AutoPDE-specific additions. We use temperature $0.2$ and \texttt{max\_tokens}$=8{,}192$. Prompts, tool schemas, and skill-library contents are documented in Appendix~\ref{app:impl} and Appendix~\ref{app:prompts}.

\paragraph{Fairness of the shared-backbone comparison.}
The four Claude-Opus-4.6 rows of Table~\ref{tab:main-results} and, separately, the four GPT-5.1 rows differ only in their scaffolding: within each backbone block, all four configurations share the language model, the benchmark prompt, the per-case wall-clock cap, and the evaluator. The Naive LLM runs use temperature $0.0$ to isolate the model's direct solver-writing ability from any sampling-induced variance; the three interactive agent scaffolds use temperature $0.2$ to allow exploration across turns (lowering it further does not change the ordering in our pilots). We do not add extra retries or repeated trials beyond each baseline's standard inference procedure: each row in Table~\ref{tab:main-results} reports one benchmark run per case.

\section{Implementation Details}
\label{app:impl}

Within PDE Agent Bench, AutoPDE is exposed through the \texttt{ustcpdeagent} wrapper. For each case, the wrapper creates an isolated workspace, instantiates the same execution environment and model adapter used by the baseline agents, and runs fully autonomously until a single \texttt{solver.py} is submitted. The orchestrator concatenates the base interactive-agent rules with a short PDE-specific protocol that asks the model to externalise its solver strategy before coding. The only AutoPDE-specific runtime tools are \texttt{get\_pde\_skill}, which retrieves a family-level numerical-method guide, and \texttt{get\_profiling\_guide}, which returns the adaptive testing protocol used before final submission.

The implementation uses a two-pass workflow. The first pass is the strategy-and-code generation pass. It must write a complete \texttt{solver.py}, execute it when possible, and revise the solver if the observed error or runtime violates the benchmark target. The second pass is a lightweight review pass with a fresh context and the same backbone. It reads the submitted solver and may apply only local repairs (e.g. missing imports, shape mismatches, output-key mistakes, or small API fixes); it is instructed not to redesign the numerical method or replace the solver strategy. This separation keeps the numerical contribution in the first pass while making the benchmark interface robust.

The PDE skill library is intentionally lightweight. Each skill is a text description rather than executable code, and covers the formulation, discretisation, stabilisation or mixed-space choices, solver/preconditioner recommendations, and common failure modes for one equation family. The skills are therefore closer to reusable numerical-method notes than to templates: the agent still has to instantiate the variational form, boundary conditions, and output format for each case. In all main experiments we use the same per-case wall-clock cap and evaluator as PDE Agent Bench, temperature $0.2$ for interactive agent rows, temperature $0.0$ for direct single-shot LLM rows, and the same backbone-specific API for all methods in a backbone block.

\section{Prompt Templates}
\label{app:prompts}

AutoPDE's prompt is organised around two compact strategy cards. We include the schema below to make the protocol reproducible without exposing task-specific examples or long library text.

\begin{lstlisting}[style=promptcard]
DIAGNOSIS
  equation_type:        poisson / heat / navier_stokes / ... / other
  spatial_dim:          1 / 2 / 3
  unknowns:             scalar / vector / vector+scalar / scalar+scalar
  coupling:             none / sequential / saddle_point
  linearity:            linear / nonlinear
  time_dependence:      steady / transient
  dominant_physics:     diffusion / convection / reaction / wave / mixed
  peclet_or_reynolds:   low / moderate / high / N/A
  solution_regularity:  smooth / boundary_layer / discontinuity / unknown
  bc_type:              all_dirichlet / mixed / periodic
  special_notes:        pressure_pinning / manufactured_solution / ...

METHOD
  spatial_method:       fem / fdm / spectral
  element_or_basis:     Lagrange_P2 / Taylor-Hood_P2P1 / upwind_fd / ...
  stabilization:        none / supg / upwind / other
  time_method:          none / backward_euler / crank_nicolson / bdf2 / rk4
  nonlinear_solver:     none / newton / fixed_point
  linear_solver:        cg / gmres / minres / direct_lu
  preconditioner:       amg / ilu / jacobi / none
  special_treatment:    pressure_pinning / continuation / splitting / none
  pde_skill:            family name when a FEM skill is applicable
\end{lstlisting}

The generator is instructed to fill \texttt{DIAGNOSIS} first, then \texttt{METHOD}, retrieve a PDE skill when the selected family has one, and only then write \texttt{solver.py}. The profiling guide is intentionally generic: run the submitted solver, print the measured error and wall time, increase mesh resolution or polynomial degree when the error is too high, simplify the solver or loosen tolerances when runtime exceeds the budget, and do not submit until the observed metrics satisfy the benchmark constraints. The reviewer prompt is restricted to local fixes and explicitly forbids changing the chosen numerical method.

\section{Case Study Trace}
\label{app:cases}

We include one condensed trace from the convection--diffusion case \texttt{convdiff\_high\_pe}. The case has $\varepsilon=10^{-2}$ and $\boldsymbol\beta=(10,10)$, so $\mathrm{Pe}=1000$ and the problem is strongly convection dominated. The trace below omits tool chatter and file I/O, keeping only the solver-strategy decisions that lead to the final code.

\begin{lstlisting}[style=promptcard]
DIAGNOSIS
  equation_type:        convection_diffusion
  spatial_dim:          2
  unknowns:             scalar
  coupling:             none
  linearity:            linear
  time_dependence:      steady
  dominant_physics:     convection / diffusion
  peclet_or_reynolds:   high
  solution_regularity:  boundary_layer
  bc_type:              all_dirichlet
  special_notes:        manufactured_solution

METHOD
  spatial_method:       fem
  element_or_basis:     Lagrange_P2
  stabilization:        supg
  time_method:          none
  nonlinear_solver:     none
  linear_solver:        gmres
  preconditioner:       ilu
  special_treatment:    none
  pde_skill:            convection_diffusion

Generated solver choices
  mesh_resolution:      N = 100
  element_degree:       p = 2
  stabilization_tau:    h / (2 |beta|)
  SUPG_term:            tau (beta . grad(v))
                        (beta . grad(u) - eps div grad(u))
  linear_solver:        GMRES + ILU

Benchmark result
  rel_L2_error:         5.85e-7
  target_error:         5.44e-4
  wall_time:            1.05 s
  target_time:          3.48 s
  final_status:         PASS
\end{lstlisting}

This trace illustrates the numerical-strategy pattern emphasized in Section~\ref{sec:analysis}. The agent first identifies the high-P\'eclet regime, then couples three decisions that must be consistent for this operator: a P2 finite-element space, SUPG stabilisation with a convection-dominated $\tau$, and a non-symmetric Krylov solver. The final solver is therefore not merely ``a code patch that happens to run'', but an implementation of an explicit strategy matched to the operator's convection-dominated structure.

\end{document}